\documentclass[12pt,a4paper]{article}
\usepackage{fullpage}
\usepackage{authblk}

\usepackage{mathrsfs}
\usepackage{amsfonts}
\usepackage{amssymb}
\usepackage{bm}
\usepackage{enumerate}
\usepackage{tikz}
\usepackage{graphicx}
\usepackage{stfloats}
\usepackage{amsmath}

\usepackage{mathrsfs}
\usepackage{amsfonts}
\usepackage{amssymb}
\usepackage{bm}
\usepackage{enumerate}
\usepackage{tikz}
\usepackage{graphicx, multirow, subfigure, txfonts}
\usepackage{stfloats}
\usepackage{amsmath}
\usepackage[colorlinks,
            linkcolor=red,
            anchorcolor=blue,
            citecolor=green
            ]{hyperref}

\newtheorem{Def}{Definition}[section]

\numberwithin{equation}{section}

\title{Facial Expression Classification Using Rotation Slepian-based Moment Invariants }

\author{Cuiming~Zou\thanks{zoucuiming2006@163.com}  }

\author{Kit Ian Kou\thanks{Corresponding author: kikou@umac.mo}}

\affil{\normalsize{Department of Mathematics, Faculty of Science and Technology, University of Macau, Macao, China}}
\date{}

\begin{document}
   \maketitle

\begin{abstract}
\normalsize

Rotation moment invariants have been of great interest in image processing and pattern recognition.
This paper presents a novel kind of rotation moment invariants based on the Slepian functions,
which were originally introduced in the method of separation of variables for Helmholtz equations. They were first proposed for time series
by Slepian and his coworkers in the 1960s.
Recent studies have shown that these functions have an good performance in local approximation compared to other approximation basis.
Motivated by the good approximation performance, we construct the Slepian-based moments and derive the rotation invariant.
We not only theoretically prove the invariance, but also discuss the experiments on real data.
The proposed rotation invariants are robust to noise and yield decent performance in facial expression classification.
\end{abstract}


\section{Introduction}
\label{sec:S1}
Moment invariants have aroused great interest in image processing and pattern recognition
in the last 50 years  \cite{TC1988}.
The first kind of moment invariants was presented by Hu \cite{H1962} in pattern recognition.
Since then, the applications have been reported in a variety of problems, such as image watermark \cite{KL2003},
gesture recognition \cite{PBTC2010}, noise analysis \cite{H1999} and color image \cite{PC1999}.
Invariance with respect to translation, rotation and scaling is required in almost all practical applications, because the article should be correctly acquainted, regardless of its position and orientation in the site and of the object-to-camera distance. Furthermore, the translation, rotation and scaling model is a ample approximation of the actual image deformation if the location is flat and almost vertical to the optical axis. Therefore, much attention has been paid to translation, rotation and scaling invariants.
While translation and scaling invariants can be derived in a instinctive way, derivation of invariants to rotation is far more complicated.

The moment $M_{mn}$ of a weighting kernel $\phi_{mn}$
and the intensity function $f$ is given by \cite{FSZ2009}
\begin{eqnarray}
M_{mn}:= \iint_{\mathbb{R}^2}\phi_{mn}(x,y)f^*(x,y)dxdy,
\end{eqnarray}
where $M_{mn}$ $(m, n =0, 1,2,3,\cdots)$ are the projections of $f$
onto the space spanned by $\{\phi_{mn} \}_{m,n=0}^{\infty}$.
Moments $M_{mn}$ can be grossly divided into non-orthogonal moments and orthogonal moments \cite{SLC2007}.
The classical non-orthogonal moments include
geometric moments, rotational moments and complex moments.
The geometric moments are projections of the image onto the space spanned by monomials, i.e., $\phi_{mn}(x,y)=x^my^n$  \cite{MR1998}.
The complex moments have the similar structure as geometric moments except for $\phi_{mn}(x,y)=(x+\mathbf{i}y)^m(x+\mathbf{i}y)^n$  \cite{RPBWS2003}.
The rotational moments are defined directly in polar coordinates, i.e., $\phi_{mn}(r,\theta)=r^me^{\mathbf{i}n\theta}$  \cite{MYP1985}.

From the theory of algebraic invariants, the non-orthogonal basis will lead to the information redundancy \cite{FSZ2009}.
In order to overcome the shortcomings of this problem, 
Teague \cite{T1980} introduced the orthogonal moments by orthogonal polynomials in 1980.
Some popular orthogonal moments are Legendre and Zernike moments \cite{Z2012}.
Legendre moments \cite{K2007} in Cartesian coordinate are generated by $\phi_{mn}(x,y):=P_m(x)P_n(y)$,
where $P_m(x)$ is the $m$th order of Legendre polynomial. The finite energy signal $f(x,y)$ can be reconstructed by its Legendre moments \cite{T1980}. While Zernike moments \cite{WH1998} were generated by $\phi_{mn}(r,\theta):=R_m(r)e^{-\mathbf{i}n\theta}$,
where $R_m(r)$ is the $m$th order Zernike radial polynomial.
Other kinds of continuous orthogonal moments have the similar structure and properties as the above moments,
such as pseude-Zernike moments \cite{XZSHL2007}, Chebyshev-Fourier moments \cite{PWS2002}, Exponent-Fourier moments\cite{HZSJ2014}.
There are some discrete orthogonal moments in the literature,
such as Tchebichef, dual Hahn, Racah moments and so on,
which can be effectively solved a lots of physical problems \cite{M2004}.

The paper proposes a novel kind of moments  on a disk.  
They are constructed as products of a radial factor (
 Slepian functions) and angular factor (harmonic function). We named it as Slepian-based moments (SMs) and used SMs to construct rotation invariants. Slepian functions are a set of functions derived by timelimiting and lowpassing, and a second timelimited operation, as originally proposed for time series by Slepian and coworkers, in the 1960s \cite{SP1961}.
They possess many remarkable properties, such as orthogonal basis of both square integrable space of finite interval and the Paley-Wiener space of bandlimited functions on the real line \cite{JJ2011, SP1961, S1964}. They are the most energy concentrated functions, among the set of bandlimited functions with a given bandwidth, on fixed space domain \cite{S1964}. Since the energy concentrated property on fixed space domain,
the Slepian basis has good performance in local approximation compared to other approximation basis, such as Fourier basis and polynomials
\cite{JJ2011}. This raises the question of whether they yield decent performance in image processing.

The aim of the article is to answer this question by applying the Slepian functions as the kernel $\phi_{mn}$ of moment $M_{mn}$. To the best of the authors' knowledge, the study of moment invariants under the Slepian functions has not been carried out.
In the present study we construct the SMs and derive the rotational invariants.
We show that the numerical computation of SMs is  $O(MN\log_2N)$ and apply these new invariants in facial expression classification.
Our experiments demonstrate that the rotational invariants are robust to image noise and
have good recognition capabilities in facial expression classification.

The paper is organized as follows.
Section \ref{sec:S2} is devoted to necessary preliminaries.
In Section \ref{sec:S3} the rotation invariants are derived by the proposed Slepian-based  moments.
In Section \ref{sec:S4}, the extended Conh-Kanade database (CK$+$) \cite{LCKSA2010} is used to test the rotation invariance and the robustness of the proposed variants to noise.
The proposed SMs invariants are also applied as features to facial expression classification. Finally, we draw a conclusion in Section \ref{sec:S5}.

\section{Preliminary}
\label{sec:S2}

In the following,  the {\it image} $f$ denotes a piecewise continuous complex-valued function defined on a compact domain of $2$D space ($D\subset \mathbb{R}^2$) \cite{FSZ2009}. The image function $g(x,y)$ in Cartesian coordinate can be changed to polar coordinate as $g(r\cos\theta,r\sin\theta)$.
To simplify the notation, denote $f(r,\theta)$ the polar coordinate of an image in the following.

\subsection{Moment in polar coordinates}
\begin{Def}
Moment $M_{mn} $ of  an image $f$ in  polar coordinates is defined by
\begin{eqnarray}
M_{mn}:= \iint_{D} R_{mn}(r)e^{-\mathbf{i}n\theta} f^*(r,\theta)rdrd\theta,
\end{eqnarray}
where $m,n$ are non-negative integers, $r=m+n$ is the order of $M_{mn} $ and $( \cdot)^*$ means the complex conjugate operation.
$(r,\theta)$ are the polar coordinates. $R_{mn}(r)$ is a radial part and $e^{-\mathbf{i}q\theta}$ is the angular factor.
The area of  $D:=\{(r,\theta): 0\leq r\leq 1, 0\leq \theta \leq 2\pi \}$. 
\end{Def}

There are many moments defined on a disk, for example,
\begin{itemize}
  \item the Zernike moments of the $m+n$-th order are defined as the radial part $R_{m}(r)$ is the Zernike polynomials;
  \item the Fourier-Mellin moments have the similarly definition as $R_{m}(r)= r^{n-2}$;
  \item  the orthogonal Fourier-Mellin moments defined as $R_{m}(r)$ is a linear combination of the factor $r^i$, $i=0,1,2...$, i.e.,
\begin{eqnarray}
R_{m}(r)=\sum_{s=0}^{m}(-1)^{m+s}\frac{(m+s+1)!}{(m-s)!s!(s+1)!)}r^s.
\end{eqnarray}
This radial functions $R_{m}(r)$ are modified by  $r^i$, $i=0,1,2...$.
\end{itemize}

Numerous types of moments with orthogonal polynomials have been described in different area of practical applications \cite{FSZ2009}.
The finite energy image $f$ can be reconstructed by these orthogonal polynomials
\begin{eqnarray}
f(r, \theta)=\sum_{m,n=0}^{\infty}M_{mn}R_{mn}(r)e^{-\mathbf{i}n\theta}.
\end{eqnarray} It minimizes the mean-square error when using only
a finite set of moments on the reconstruction \cite{FSZ2009}.

\subsection{Slepian-based moment}

Similar to the definition of moment in polar coordinates, such as Zernike moments, Fourier-Mellin moments, orthogonal Fourier-Mellin moments.
We will change the radial part to a special function i.e., Slepian functions. The moments we defined by this functions is called  Slepian-based moments.
Now we first introduce the Slepian functions in continuous form and discrete form.

\subsection{Slepian functions in continuous form}

The Slepian functions, originated from the context of separation of variables for the Helmhotz equation in spheroidal coordinates, have been extensively used for a variety of physical and engineering applications \cite{FSZ2009, S1964}.
It is remarkable that the Slepian functions $\{\psi_{mn}\}_{m,n=0}^{\infty}$ \cite{S1964} are also the eigenfunctions of the integral equations
\begin{eqnarray}\label{eq:2Slepianintegral}
\lambda_{m}\psi_{m}(x)=\int_{-1}^{1} \psi_{m}(u)e^{\mathbf{i}cxu}du,
\end{eqnarray}
where the integral domain  is symmetric and $c$ is a given positive number and $\lambda_{m}$
is the real-valued eigenvalue corresponding to $\psi_{m}$ in Eq. (\ref{eq:2Slepianintegral}).

This kind of functions have many good properties, the most important properties are list as follows:
given a real $c>0$, there are a countably infinite set of $\{\psi_{n}(x)\}_{n=0}^{\infty}$
and their corresponding eigenvalues $\{ \lambda_{n} \}_{n=0}^{\infty}$
\begin{itemize}
  \item
   The eigenvalues $\lambda_n$'s are real and monotonically decreasing in $(0,1)$:
\begin{eqnarray}\label{eigenvaluerelation}
\lambda_{0} \geq\lambda_{1} \geq\lambda_{2} \geq \cdots,
\end{eqnarray}
and such that $\lim_{n\rightarrow \infty} \lambda_{n}=0$;

  \item The $\{ \psi_{n}(x) \}_{n=0}^{\infty}$ are orthogonal in $[-1,1]$:
\begin{eqnarray}
\int_{-1}^1 \psi_{n}(x) \overline{\psi_{m}(x)} dx= \lambda_{n}\delta_{mn};
\end{eqnarray}

\item The $\{ \psi_{n} \}_{n=0}^{\infty}$ are orthonormal in $\mathbb{R}^2$:
\begin{eqnarray}
\int_{\mathbb{R}^2} \psi_{n}(x) \overline{\psi_{m}(x)} dx= \delta_{mn}.
\end{eqnarray}
\end{itemize}

The Slepian functions are also an orthogonal basis of both integrable in the finite interval
and the Paley-Wiener space of bandlimited functions on the real line \cite{SP1961, S1964, JJ2011}.
The Slepian functions have been proved as the most energy concentrated functions,
among the set of bandlimited functions with a given bandwidth, on fixed space domain \cite{S1964}.
Since the energy concentrated property on fixed space domain,
the Slepian basis has good performance in local approximation compared to other approximation basis,
such as Fourier basis and polynomials \cite{JJ2011}.
This raises the question of whether they yield decent performance in image processing.

\subsection{Slepian functions in discrete form}
Recently, there has been a growing interest in developing numerical methods using Slepian functions as basis functions.
The discrete Slepian functions are developed by Slepian at first \cite{S1978}, which is also called the discrete prolate spheroidal sequences (DPSS).
For given $N$ the number of length of the sequences and $W$ the bandwidth of the sequences, the DPSS ${v_n^{(k)}(N,W)}$ is defined as
the solution to the system of equations
\begin{eqnarray}
\sum_{m=0}^{N-1}\frac{\sin 2\pi(n-m)}{\pi(n-m)}{v_m^{(k)}(N,W)}
=\lambda_k(N,W){v_n^{(k)}(N,W)},
\end{eqnarray}
where $k=0,1,2,...,N-1$, $\lambda_k(N,W)$
is the eigenvalue for the DPSS ${v_n^{(k)}(N,W)}$.
The DPSS are also have the double orthogonal properties
\begin{eqnarray}
\sum_{m=0}^{N-1}{v_m^{(s)}(N,W)}{v_m^{(t)}(N,W)}
=\lambda_s(N,W)\sum_{m=0}^{N-1}{v_m^{(s)}(N,W)}{v_m^{(t)}(N,W)}
=\delta_{st},
\end{eqnarray}
where $\delta_{st}=0$ for $s\neq t$ and $\delta_{st}=1$ if $s=t$.
Here, $s,t=0,1,2,...,N-1$.

Denote
\begin{eqnarray}
f_k(N,W,u):=\epsilon_k\sum_{m=0}^{N-1}{v_m^{(k)}(N,W)}
e^{-\mathbf{i}\pi (N-1-2m)u},
\end{eqnarray}
where $\epsilon_k=1$ for $k$ even and $\epsilon_k=0$ for $k$ odd.
and the coefficient ${v_m^{(k)}(N,W)}$ has
\begin{eqnarray}
{v_m^{(k)}(N,W)}=\frac{1}{\epsilon_k}\int_{-\frac{1}{2}}^{\frac{1}{2}}
 f_k(N,W,u)e^{\mathbf{i}\pi (N-1-2m)u}du.
\end{eqnarray}
We may also obtain that
\begin{eqnarray}
{v_m^{(k)}(N,W)}=\frac{1}{\epsilon_k \lambda_k(N,W) }
\int_{-W}^{W} f_k(N,W,u)e^{\mathbf{i}\pi (N-1-2m)u}du.
\end{eqnarray}
Let $W=1$, the above formulary is already the Slepian functions in
the integer points, i.e., $x=N-1-2m$ in Eq. (\ref{eq:2Slepianintegral}).

\subsection{Slepian-based moments}
Since we have known the Slepian functions in continuous and discrete forms,
we can apply them in moment theory.
To the best of the authors knowledge, the study of moment invariants under the Slepian functions has not been carried out.
We are now ready to construct a new kind of moment related to this Slepian functions, which is named as Slepian-based moment.
The definition is similar to the definition of the moment in polar coordinates, while the radial function changes to the Slepian functions.

\begin{Def} 
The Slepian-based moments (SMs) of an image $f(r,\theta)$ is defined by
\begin{eqnarray}\label{eq:SM}
S_{mn}:= \iint_{D} \psi_{m}(r)e^{-\mathbf{i}n\theta} f^*(r,\theta)rdrd\theta,
\end{eqnarray}
where $m,n$ are non-negative integers and $r=m+n$ is the order of $S_{mn}$.
The radial functions $\{\psi_{m}\}_{m=0}^{\infty}$ are the Slepian functions satisfying Eq. (\ref{eq:2Slepianintegral}).
\end{Def}
The bandlimited functions $f(r,\theta)$  can be represented by the Slepian series \cite{S1964}
\begin{eqnarray}
f(r,\theta)=\sum_{m=0}^{\infty}\sum_{n=0}^{\infty}S_{mn}\psi_{m}(r)e^{-\mathbf{i}n\theta}.
\end{eqnarray}
Moreover, it can also be approximated well by linear combinations of the Slepian functions of order $M+N$ \cite{S1964}.
\begin{eqnarray}
\widetilde{f}(r,\theta)=\sum_{m=0}^{M}\sum_{n=0}^{N}S_{mn}\psi_{m}(r)e^{-\mathbf{i}n\theta}.
\end{eqnarray}

\section{Rotation invariants}\label{sec:S3}

In this section, the use of SMs in in polar coordinates on $D=\{(r,\theta): 0\leq r\leq 1, 0\leq \theta \leq 2\pi \}$
in the capacity of rotation invariants is discussed.
For any image function $f(r,\theta)$ in polar coordinate.

\subsection{Rotation invariance}

Let $f'(r,\theta)$ be a rotated version of $f(r,\theta)$ with a rotation angle $\alpha$, i.e., $f'(r,\theta)=f(r,\theta+\alpha)$.
The $S'_{mn}$ of $f'(r,\theta)$ and $S_{mn}$ of $f(r,\theta)$ have relations
\begin{eqnarray*}\label{Rotation}
S'_{mn}&=&\int_{0}^{1}\int_{0}^{2\pi}\psi_{m}(r)e^{-\mathbf{i}n\theta}f^*(r,\theta+\alpha)\sqrt{r}d\theta dr\\
&=&e^{\mathbf{i}m\alpha}\int_{0}^{1}\int_{0}^{2\pi}\psi_{mn}(r)e^{-\mathbf{i}m\theta}f^*(r,\theta)\sqrt{r}d\theta dr\\
&=&e^{\mathbf{i}m\alpha}S_{mn}.
\end{eqnarray*}
The above equation implies rotation invariance of the $S_{mn}$ while the phase is shifted by $\alpha$.
Taking absolute value on both sides of $S'_{mn}=e^{\mathbf{i}m\alpha}S_{mn}$, we have
\begin{eqnarray}
|S'_{mn}|=|e^{\mathbf{i}m\alpha}S_{mn}|=|S_{mn}|=:\Phi_{mn}.
\end{eqnarray}
Thus, $\Phi_{mn}$ is a rotation invariant induced by $S_{mn}$.
The rotation invariance is the most important invariance for the moment theory.
Since we find these rotation invariants, we can use them to the practical applications.
In the following, we will show the numerical computation for this kind of moments.

\subsection{Numerical computation}
Considering a discrete $M \times N$ image $g(M, N)$, we first change the Cartesian coordinates to polar coordinates \cite{XPL2007}.
For a  polar coordinates image $f(r,\theta)$, calculate the discrete form of Eq. (\ref{eq:SM}) in polar coordinates as follows
\begin{eqnarray}\label{eq:DisSM}
{\cal S}_{mn} &:=&
\sum_{r=0}^{1}\sum_{\theta=0}^{2\pi}\psi_{m}(r)
e^{-\mathbf{i}n\theta}f^*(r,\theta)r\Delta\theta\Delta r\\
&=& \sum_{r=0}^{1}\psi_{m}(r) r
\left(\sum_{\theta=0}^{2\pi}e^{-\mathbf{i}m\theta}f^*(r,\theta)\Delta\theta \right)\Delta r, \nonumber
\end{eqnarray}
where the expression inside the parenthesis can be considered as a discrete form of one-dimensional Fourier transform in $\theta$.
Therefore, we can use the FFT in matlab to obtain fast computation.
As for numerical computation of the function $\psi_{m}(r)$, please refer to  the programs of DPSS in \cite{S1978}.
The total computing complexity of $S_{mn}$ is $O(MN\log_2N)$.

\section{Experimental results}
\label{sec:S4}
This section is devoted to verify the performance of SMs and SMs rotation invariants in  facial expression classification,
which is a kind of problem in facial classification \cite{CKW2016}.
Specifically, the extended Conh Kanade database (CK$+$) \cite{LCKSA2010} is used for the experiments.

\begin{figure}
  \centering
  \includegraphics[height=2.5cm]{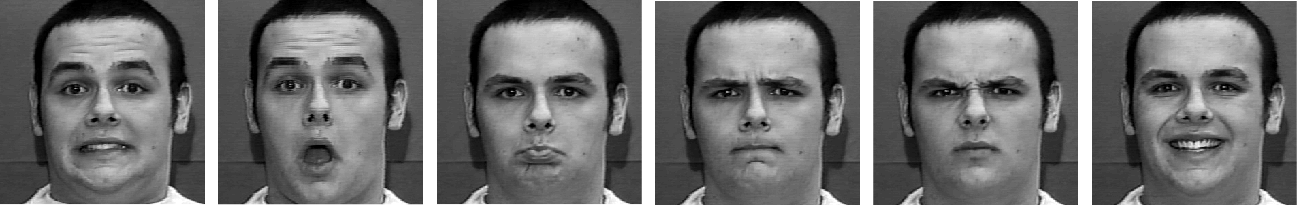}
 \caption{Examples in the CK$+$ database. Facial images of $6$ emotions:  Fear, Surprise, Sadness, Disgust, Anger, and Happy. }
\label{fig:example}
\end{figure}
\subsection{Image data}
The CK$+$ is a kind of database to detect individual facial expressions \cite{LCKSA2010}.
We consider the posed emotion-specified expressions, which have $593$ image sequences from $123$ subjects.
These sequences have a nominal emotion label based on the subject's impression
of each of the basic emotion categories: Fear, Surprise, Sadness, Disgust, Anger, and Happy.
Examples of the CK$+$ are given in Figure~\ref{fig:example}.

\begin{figure}
  \centering
  \includegraphics[height=5.0cm]{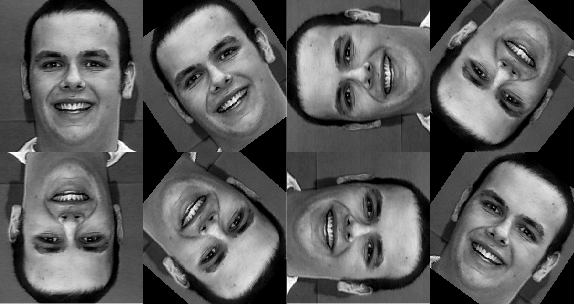}
 \caption{Examples of an image in $8$ orientations:
 $0^\circ$, $  35^\circ$, $  90^\circ$, $140^\circ$, $180^\circ$, $230^\circ$, $270^\circ$, and $325^\circ$.}
 \label{fig:rotationclearimage}
\end{figure}

\begin{table*}
\caption{$\Phi_{mn}$ and  standard deviations  for the same image in $8$ orientations in Figure~\ref{fig:rotationclearimage}.}
\begin{center}
\hspace{-1.0cm}
\begin{tabular}{c|c|c|c|c|c|c|c|c|c|c}
\hline\cline{1-11}
$\alpha$ &$~\Phi_{1,1}$  &$~\Phi_{1,2}$ &$~\Phi_{2,1}$ &$~\Phi_{2,2}$ &$~\Phi_{2,3}$ &$~\Phi_{3,2}$ &$~\Phi_{3,4}$ &$~\Phi_{4,1}$ &$~\Phi_{4,3}$ &$~\Phi_{4,5}$\\
\hline
$0^\circ$&0.0517&0.0764&0.0733&0.0955&0.0476&0.1202&0.1559&0.2366&0.3316&0.0881\\
$  35^\circ$&0.0612&0.0750&0.0770&0.0933&0.0461&0.1185&0.1514&0.2510&0.3474&0.0984\\
$  90^\circ$&0.0724&0.0742&0.0855&0.0926&0.0479&0.1175&0.1553&0.2280&0.3295&0.0864\\
$140^\circ$&0.0612&0.0754&0.0764&0.0929&0.0455&0.1172&0.1429&0.2411&0.3277&0.0959\\
$180^\circ$&0.0724&0.0740&0.0815&0.0914&0.0446&0.1172&0.1471&0.2672&0.3645&0.1085\\
$230^\circ$&0.0612&0.0748&0.0766&0.0926&0.0458&0.1175&0.1426&0.2407&0.3273&0.0957\\
$270^\circ$&0.0519&0.0764&0.0688&0.0942&0.0444&0.1196&0.1478&0.2735&0.3659&0.1117\\
$325^\circ$&0.0611&0.0753&0.0768&0.0934&0.0462&0.1186&0.1514&0.2509&0.3489&0.0978\\
\hline
$std.$&\textbf{0.0073}&\textbf{0.0008}&\textbf{0.0047}&\textbf{0.0011}
&\textbf{0.0012}&\textbf{0.0011}&\textbf{0.0047}&\textbf{0.0144}&\textbf{0.0151}&\textbf{0.0082}\\
\hline\cline{1-11}
\end{tabular}
\end{center}
\label{table1}
\end{table*}

\subsection{Rotation invariance and noise}
In Section \ref{sec:S3}, we have proved that $\Phi_{mn}$ are rotation invariants based on $S_{mn}$ for different $m,n$.
In this part, we will compute the $\Phi_{mn}$ for the same image
with different orientations to validate the rotation invariance of $\Phi_{mn}$ numerically.
The rotation transform with the angle $\theta$ for an image is as follows
\begin{eqnarray*}
\left(\begin{array}{ll}x'\\y'\end{array}\right)=
\left(\begin{array}{ll}\cos\theta&-\sin\theta\\\sin\theta&\cos\theta\end{array}\right)
\left(\begin{array}{ll}x\\y\end{array}\right).
\end{eqnarray*}
Figure~\ref{fig:rotationclearimage} shows an image with $8$ different orientations.
Since the different rotation will change the size of images,
we just show the rotational images with the same size as the original one.
In Table  \ref{table1} we show $10$ values of $\Phi_{mn}$ with $m=1,2,3,4$ and $n=1,2,3,4,5$ for every rotated image.
The last row of Table  \ref{table1} is the standard deviation (\emph{std.}) of $\Phi_{mn}$ in different orientations.
Since all of the \emph{std.} for different $\Phi_{mn}$ are very small, it shows that the value of $\Phi_{mn}$ is stable.
This result is consistent with the proposed theory of the rotational invariance of $\Phi_{mn}$.

\begin{figure}
  \centering
  \includegraphics[height=5.0cm]{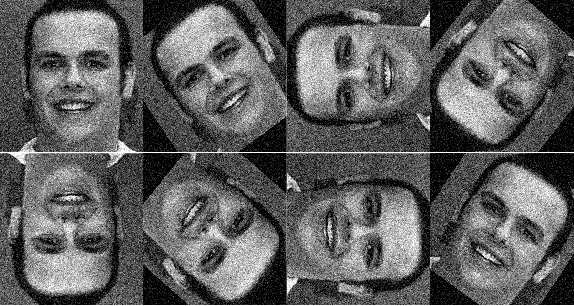}
 \caption{Examples of an image in $8$ orientations with $30$db Gaussian noise.:
 $0^\circ$, $  35^\circ$, $  90^\circ$, $140^\circ$, $180^\circ$, $230^\circ$, $270^\circ$, and $325^\circ$.}
 \label{fig:rotationnoiseimage}
\end{figure}

Now we evaluate the robustness of the $\Phi_{mn}$ to noise \cite{HC1995}.
As an example in Figure~\ref{fig:rotationnoiseimage} we show the mostly widely used Gaussian noise to the same images in Figure~\ref{fig:rotationclearimage}.
In Table  \ref{table2} we experiment the values of $\Phi_{mn}$ for the images in Figure~\ref{fig:rotationclearimage}
with $30$db Gaussian noise for $m=1,2,3,4$ and $n=1,2,3,4,5$.
Even all of the \emph{std.} in Table  \ref{table2} for different $\Phi_{mn}$
is slight larger than the \emph{std.} in Table  \ref{table1}, but they are also very small.
It shows that for the same image with Gaussian noise, the value of $\Phi_{mn}$s are stable.
We can now find that the value of $\Phi_{mn}$ of same image with noise
are basically the same as the value of $\Phi_{mn}$ in Table  \ref{table1}.
That is, the noise has little influence to the $\Phi_{mn}$.

\begin{table*}
\caption{$\Phi_{mn}$ and  standard deviations  for the images in Figure~\ref{fig:rotationclearimage} with $30$db Gaussian noise.}
\begin{center}
\hspace{-1.0cm}
\begin{tabular}{c|c|c|c|c|c|c|c|c|c|c}
\hline\cline{1-11}
$\alpha$ &$~\Phi_{1,1}$  &$~\Phi_{1,2}$ &$~\Phi_{2,1}$ &$~\Phi_{2,2}$ &$~\Phi_{2,3}$ &$~\Phi_{3,2}$ &$~\Phi_{3,4}$ &$~\Phi_{4,1}$ &$~\Phi_{4,3}$ &$~\Phi_{4,5}$\\
\hline
$  0^\circ$&0.0536&0.0782&0.0720&0.0973&0.0456&0.1202&0.1568&0.2414&0.3266&0.0884\\
$  35^\circ$&0.0632&0.0718&0.0779&0.0921&0.0467&0.1207&0.1530&0.2431&0.3516&0.1039\\
$  90^\circ$&0.0769&0.0753&0.0869&0.0926&0.0488&0.1171&0.1556&0.2305&0.3393&0.0856\\
$140^\circ$&0.0667&0.0748&0.0761&0.0956&0.0482&0.1203&0.1445&0.2360&0.3263&0.0970\\
$180^\circ$&0.0733&0.0729&0.0788&0.0903&0.0438&0.1150&0.1495&0.2766&0.3674&0.1053\\
$230^\circ$&0.0605&0.0758&0.0762&0.0909&0.0462&0.1172&0.1433&0.2348&0.3250&0.0908\\
$270^\circ$&0.0506&0.0760&0.0695&0.0942&0.0459&0.1201&0.1448&0.2706&0.3693&0.1137\\
$325^\circ$&0.0594&0.0705&0.0750&0.0939&0.0467&0.1200&0.1493&0.2473&0.3402&0.0996\\
\hline
$std.$&\textbf{0.0085}&\textbf{0.0024}&\textbf{0.0048}&\textbf{0.0022}&\textbf{0.0015}
&\textbf{0.0020}&\textbf{0.0048}&\textbf{0.0159}&\textbf{0.0168}&\textbf{0.0089}\\
\hline\cline{1-11}
\end{tabular}
\end{center}
\label{table2}
\end{table*}

\subsection{Facial expression classification}
At last, we aim to apply the $\Phi_{mn}$ with different $m,n$ as feature vectors to the problem of emotion-specified expression classification.
We use $\Phi_{mn}$ as features to classify the different facial expression for the same subject.

\begin{figure}[!t]
  \centering
     \includegraphics[height=6.5cm]{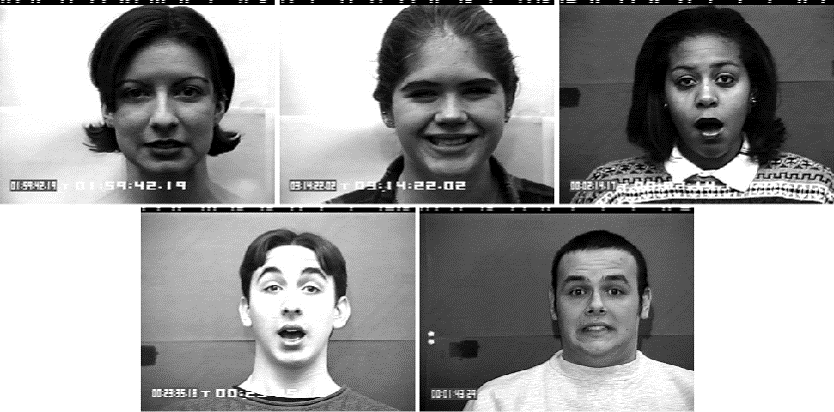}
 \caption{Examples of the 5 subjects in CK$+$ database used for classification,
 from left to right: S010, S011, S022, S037, and S050.}
  \label{fig:5subjects}
\end{figure}

The support vector machine (SVM) is applied to classify the same subject
with six different emotion-specified expressions.
The SVM was proposed by Vapnik \cite{WMCPV2000} as a very effective method for multi-class classification.
Here, we only use the first $10$ Slepian functions to construct
a feature vector with $100$ elements of the images.
We choose $5$ sequences in CK$+$ database in Figure~\ref{fig:5subjects},
which are in the files of S010, S011, S022, S037, and S050.
Each sequences has been marked  by $6$ kinds of emotion-specified expressions as Figure~\ref{fig:example},
i.e., fear: $1$, surprise: $2$, sadness: $3$, disgust: $4$, anger: $5$, and happy: $6$.
For each expression we random choose $8$ images in our experiments,
which means we have $6*8$ images for each  subject.
Then we use a part of them for training and the rest images
for testing and get the percentage of classification each time.
Since we randomly choose a part of them to get the percentage of accuracy,
we repeat $10$ times for each subject and we take
the average number for classification in Table  \ref{table3}.
For $50\%$ training means, for the same subject with $48$ images, $24$ images are using to train in the SVM.
\begin{table}
\caption{Classification accuracy for 5 subjects with 6 different facial expressions
using different percent of training images.}
\begin{center}
\begin{tabular}
{c|cccccc}
\hline\cline{1-7}
$p\%$ &~~&S010&S011&S022&S037&S050 \\
\hline
\multirow{2}{*}{$20$}&$mean$&0.6590&0.3872&0.4103&0.4795&0.4615   \\
                                     &$std.$~  &0.2271&0.1370&0.2033&0.1817&0.1781    \\
\cline{1-7}
\multirow{2}{*}{$30$}&$mean$&0.9588&0.7588&0.8118&0.9735&0.9382    \\
                                     &$std.$~  &0.0397&0.1453&0.0638&0.0352&0.0596     \\
\cline{1-7}
\multirow{2}{*}{$40$}&$mean$&0.9897&0.7966&0.8793&0.9690&0.9793     \\
                                     &$std.$~  &0.0233&0.0550&0.0592&0.0343&0.0291     \\
\cline{1-7}
\multirow{2}{*}{$50$}&$mean$&\textbf{0.9833}&\textbf{0.8875}&\textbf{0.8833}&\textbf{0.9833}&\textbf{0.9958}     \\
                                     &$std.$~  &0.0291&0.0483&0.0430&0.0215&0.0132    \\
\hline\cline{1-7}
\end{tabular}
\end{center}
\label{table3}
\end{table}
From Table  \ref{table3}, we find that only  $30\%$ images are needed to train SVM and
the percentage can be reached very high.
At the same time, we must take care to the fact that only $10$ Slepian functions
used to construct the feature vectors for each subject.
Thanks to the theory of Slepian functions \cite{SP1961, S1964, S1978},
only the lower order Slepian-based moments are needed in our experiments
and they already contain the mostly information for the emotion-specified image.

\section{Conclusion}
\label{sec:S5}
This paper develops a novel kind of moments based on the Slepian functions.
Motivated by the excellent properties of the Slepian functions, we define the Slepian-based moments and the corresponding moment rotation invariants.
We notice that the study of moments combining with Slepian functions has not been carried out in the literature.
These rotational invariants are tested as features to classify the emotion-specified expressions in the CK$+$ database and
the experimental results give a positive result in terms of the accuracy of classification.
The current study only shows the moment invariants with respect to rotation suitable for emotion-specified recognition.
The construction of the generalized invariant basis to affine and projective transforms will be considered in the upcoming paper.

\end{document}